\documentclass[runningheads]{llncs}

\usepackage{eccv}

\usepackage{eccvabbrv}

\usepackage{graphicx}
\usepackage{booktabs}
\usepackage{placeins}

\usepackage[accsupp]{axessibility}  

\usepackage{hyperref}

\usepackage{orcidlink}
\usepackage{enumitem}
\usepackage{multirow}

\usepackage[capitalize]{cleveref}
\crefname{section}{Sec.}{Secs.}
\Crefname{section}{Section}{Sections}
\Crefname{table}{Table}{Tables}
\crefname{table}{Tab.}{Tabs.}

\usepackage{pifont}
\definecolor{dgreen}{rgb}{0.0,0.6,0.0}
\newcommand{\cmark}{\textcolor{dgreen}{\ding{51}}}
\newcommand{\xmark}{\textcolor{red}{\ding{55}}}

\newcommand{\proposed}{ActionVOS\xspace}

\begin{document}

\title{ActionVOS: Actions as Prompts for \\ Video Object Segmentation} 


\author{Liangyang Ouyang\inst{1}\orcidlink{0009-0000-0733-2858} \and
Ruicong Liu\inst{1}\orcidlink{0000-0002-8460-8763} \and
Yifei Huang\inst{1}$^*$\orcidlink{0000-0001-8067-6227} \and \\ 
Ryosuke Furuta\inst{1}\orcidlink{0000-0003-1441-889X} \and
Yoichi Sato\inst{1}\orcidlink{0000-0003-0097-4537}}

\authorrunning{L.~Ouyang et al.}

\institute{The University of Tokyo \\
\email{\{oyly, lruicong, hyf, furuta, ysato\}@iis.u-tokyo.ac.jp}
}

\maketitle
\let\thefootnote\relax\footnotetext{ $^*$Corresponding author.}

\begin{abstract}
Delving into the realm of egocentric vision, the advancement of referring video object segmentation (RVOS) stands as pivotal in understanding human activities.
However, existing RVOS task primarily relies on static attributes such as object names to segment target objects, posing challenges in distinguishing target objects from background objects and in identifying objects undergoing state changes.
To address these problems, this work proposes a novel action-aware RVOS setting called \proposed, aiming at segmenting only active objects in egocentric videos using human actions as a key language prompt.
This is because human actions precisely describe the behavior of humans, thereby helping to identify the objects truly involved in the interaction and to understand possible state changes.
We also build a method tailored to work under this specific setting.
Specifically, we develop an action-aware labeling module with an efficient action-guided focal loss. 
Such designs enable ActionVOS model to prioritize active objects with existing readily-available annotations.
Experimental results on VISOR dataset reveal that \proposed significantly reduces the mis-segmentation of inactive objects, confirming that actions help the \proposed model understand objects' involvement. 
Further evaluations on VOST and VSCOS datasets show that the novel ActionVOS setting enhances segmentation performance when encountering challenging circumstances involving object state changes.
We will make our implementation available at \url{https://github.com/ut-vision/ActionVOS}.
\end{abstract}

\section{Introduction}
\label{sec:intro}
Exploring the domain of egocentric vision (first-person perspective), the development of Referring Video Object Segmentation (RVOS) is critical for comprehending human activities. 
RVOS aims at segmenting target objects using natural language expressions, serving as a foundation for machines to have a comprehensive understanding of visual-language and temporal information. 
By integrating various modalities, RVOS paves the way for groundbreaking applications in egocentric contexts, such as text-directed object identification and real-time object tracking in videos. 
This has been exemplified in recent studies, including referring expression comprehension \cite{kurita2023refego,qi2020reverie}, active object localization \cite{wu2023localizing,Zhang_2023_ICCV} and intention-driven visual grounding \cite{wang2024beyond,lee2023determinet}. 
As highlighted by recent works \cite{damen2022rescaling,darkhalil2022epic,grauman2022ego4d,lin2022egocentric,zhu2023egoobjects}, advancements in egocentric applications have led to a surge in data related to egocentric interactions. This has subsequently increased the demand for RVOS from egocentric perspectives.

\begin{figure}[tb]
  \centering
  \includegraphics[width=\linewidth]{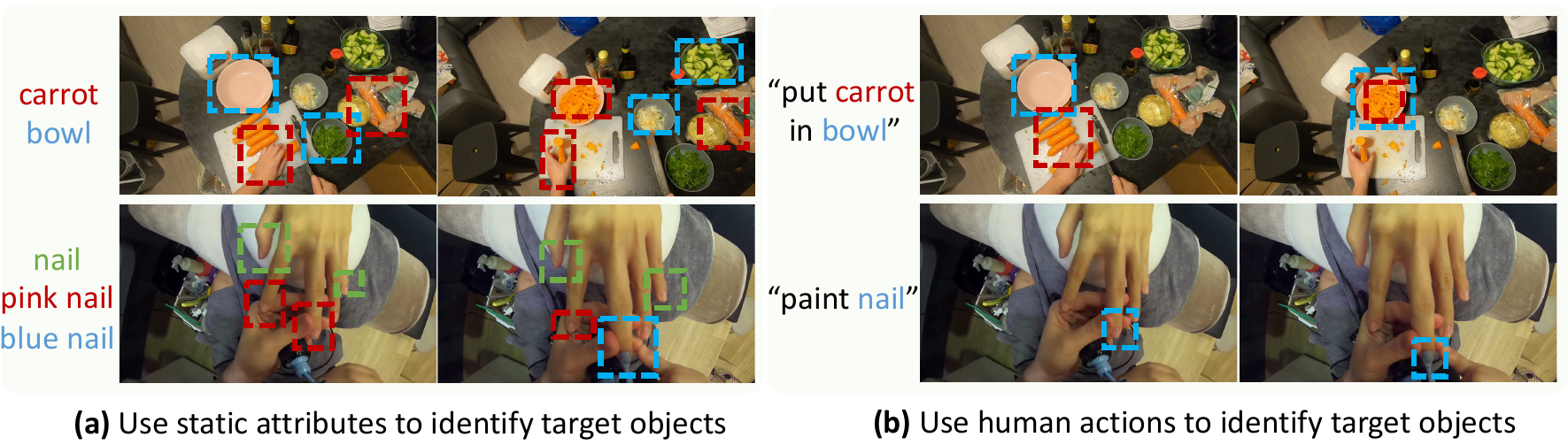}
  \caption{Human actions as language prompts help to identify active objects.}
  \label{fig:motivation}
  \vspace{-5mm}
\end{figure}

In the field of RVOS, existing benchmarks \cite{gavrilyuk2018actor,khoreva2018video,seo2020urvos} primarily rely on static attributes, \eg, object names and colors, to describe target objects in the video.
In simple scenarios \cite{kazemzadeh2014referitgame,yu2016modeling}, such static attributes are adequate to identify the target objects.
However, when scenarios become complex, these static attributes fall short in accurately identifying target objects, such as when similar redundant objects coexist or the object state is changing. 
\cref{fig:motivation} (a) illustrates two failure cases of static attributes. 
In the ``carrot\&bowl'' example, static attributes identify redundant and inactive ``carrot\&bowl''. In the case of ``nail'', static attributes fail to identify the nail painted from pink to blue.

To address these problems, we employ human actions as a substantial cue for identifying target objects.
This is because human actions, as a strong language prompt, precisely describe the behavior of humans. Such action prompts aid in identifying objects truly involved in interactions and comprehending potential object state changes. As illustrated in \cref{fig:motivation} (b), when provided with action prompt ``put carrot in bowl'', the specific carrots and bowl involved in ``put'' action are accurately identified. Similarly, the specific nail being painted is also correctly identified with the ``paint nail'' prompt. 
Therefore, action prompts significantly resolve ambiguity arising from redundant instances and object state changes.

In this work, we propose ActionVOS, a novel action-aware setting for RVOS, segmenting active objects in egocentric videos using action prompts. 
As shown in \cref{fig:input-output}, unlike conventional RVOS settings, ActionVOS incorporates an additional language prompt of action narrations.
Guided by such action prompts, ActionVOS only segment active objects involved in interactions, regardless of redundancy or state changes. 

Unfortunately, existing video object segmentation datasets \cite{darkhalil2022epic,tokmakov2023breaking,yu2023video} lack annotations of identifying active objects, \ie, whether or not they are involved in interactions.
During training, this limitation leads \proposed to face difficulty in obtaining annotations that classify whether an object is active or not.
To address this issue, we propose an action-aware labeling module to generate pseudo-labels from existing readily-available annotations, including action narrations \cite{damen2018scaling,damen2022rescaling,grauman2022ego4d}, semantic segmentation \cite{darkhalil2022epic,tokmakov2023breaking,yu2023video}, and hand-object segmentation \cite{darkhalil2022epic}.
This module enables \proposed model to obtain training data regarding the objects' involvement in actions without manually annotating their participation.
In addition, we design an effective action-guided focal loss working with the action-aware labeling module.
This proposed loss reduces the impacts of false positives in generated pseudo-labels, prioritizing the truly active objects.

We evaluate our method on three video object segmentation datasets VISOR \cite{darkhalil2022epic}, VOST \cite{tokmakov2023breaking} and VSCOS \cite{yu2023video}. Comparing with conventional RVOS setting, ActionVOS significantly reduces the mis-segmentation of inactive objects on VISOR dataset, with a 35.6\% mIoU reduction of inactive objects. Evaluation on VOST and VSCOS datasets indicates that the ActionVOS setting enhances the segmentation of objects undergoing state changes, by achieving a 3.0\% mIoU increase of state-changed objects. These results confirm that action prompts help ActionVOS model focus on active objects and enhance the understanding of state changing.

The main contributions of this work are as follows:
\vspace{-1mm}
\begin{itemize}
  \item We propose a novel action-aware setting for referring video object segmentation, ActionVOS. This setting segments active objects in egocentric videos by employing action narrations as an additional language prompt.
  \item We develop an action-aware labeling module and an action-guided focal loss for ActionVOS. This design enables ActionVOS models to segment active objects with existing readily-available annotations.
  \item Extensive evaluation results show that ActionVOS significantly reduces the mis-segmentation of inactive objects, and enhance the segmentation of state-changed objects.
\end{itemize}

\section{Related Works}
\label{sec:relatedworks}

\subsection{Referring Expression Comprehension}
Referring expression comprehension (REC) aims to localize target objects described by a referring expression in natural language.
Established REC benchmarks \cite{kazemzadeh2014referitgame,mao2016generation,hu2016segmentation,yu2016modeling,de2017guesswhat,wu2020phrasecut,liu2019clevr} and REC methods \cite{yu2018mattnet,kamath2021mdetr,wang2022ofa,luddecke2022image,zhang2022glipv2,li2022grounded,liu2023polyformer,wang2023one,yan2023universal} contribute to this fundamental yet challenging task. A new benchmark GREC \cite{he2023grec,liu2023gres} introduces generalized referring expression comprehension, extending REC by permitting expressions to describe any number of target objects.

In addition to REC in images, there has been a growing interest in video-based REC \cite{chen2019weakly,yamaguchi2017spatio,li2017tracking,deruyttere2019talk2car,wang2021towards,wu2023referring}, which requires both temporal and spatial localization of text-referred objects in video frames.
Recent works \cite{kurita2023refego,wu2023localizing,Zhang_2023_ICCV} introduce REC to track and localize active objects in egocentric videos. However, in these works, the number of target active objects is typically limited to one or two in each video. In this work, we not only extend localization to segmentation, but also aim to identify a broader range of active objects, \eg, hands, tools, containers and other entities.

\subsection{Referring Video Object Segmentation}

Referring video object segmentation (RVOS) aims to segment the target object indicated by a given expression across the entire video clip. 
Conventional RVOS datasets \cite{perazzi2016benchmark,xu2018youtube,gavrilyuk2018actor,khoreva2018video,seo2020urvos} are constructed by adding language expressions to existing video object segmentation datasets. 
These datasets often provide an expression for a single object, which usually describes the static attributes of the target object. A recent dataset MeViS \cite{ding2023mevis} focuses on segmenting objects in video content based on a sentence describing their motions.
Existing RVOS methods \cite{seo2020urvos,botach2022end,wu2022language,ding2022vlt,wu2023segment,li2023robust,cheng2023tracking,yan2023universal,wang2023seggpt,mei2024slvp} employ various approaches to address the RVOS task.
Among these works, SLVP \cite{mei2024slvp} is the first to adopt RVOS to VISOR \cite{darkhalil2022epic} dataset. Comparing to SLVP, our work incorporates an additional action narration in the language prompt to describe and segment only active objects in egocentric videos.

\subsection{Action-object Relation}
The relations between human actions and objects have been extensively studied over time.
Previous works \cite{cai2016understanding,shan2020understanding,zhang2022fine,darkhalil2022epic,fu2022sequential,Zhang_2023_ICCV,Higgins_2023_CVPR,Liu_2024_CVPR} focus on hand-object interactions as a basic for understanding active objects.
Besides hand-objects, many works have introduced different representations to model action-object relations across various applications, such as graphical models \cite{gupta2007objects}, object-action complexes (OAC) \cite{kruger2011object}, object affordances \cite{kjellstrom2011visual}, action-objects \cite{bertasius2017first}, active entities \cite{darkhalil2022epic}, objects undergoing change with tools \cite{wu2023localizing}, and action scene graphs \cite{rodin2023action}. 
In comparison to prior works, our work broadens the range of active objects in \cref{sec:task}. This includes not only objects described by action narrations but also treats hands, hand tools, containers, and contents as active objects in human actions.

\section{Problem Setting}
\label{sec:task} 

\textbf{Input.} ActionVOS task is a Referring Video Object Segmentation (RVOS) task focused on active objects involved in human actions. Its input contains three parts: 1) A video clip $\mathcal{V} = \{V_t\}^T_{t=1}$, where $V_t \in \mathbb{R}^{H\times W\times 3}$ is an RGB image from $T$ frames of the video clip. $H$ and $W$ stand for height and width, respectively. 2) An action narration $\mathcal{A}$, which describes the human action in $\mathcal{V}$. 3) A set of $N$ object names $\mathcal{O} = \{O_i\}^N_{i=1}$, where $O_i$ is a noun of an object. Note that $N$ is arbitrary and any object name can be in $\mathcal{O}$, making ActionVOS an open-vocabulary setting. 

\noindent \textbf{Output.} ActionVOS aims to predict $T$ segmentation masks $\mathcal{M} = \{M_t\}^T_{t=1}$, where $M_t \in \mathbb{R}^{N\times H\times W}$ for $N$ objects. We use $M_t(O_i)$ to represent the binary segmentation mask for $O_i$ in frame $t$, where each pixel belongs to one single object or background. 

\begin{figure}[tb]
  \centering
  \includegraphics[width=\linewidth]{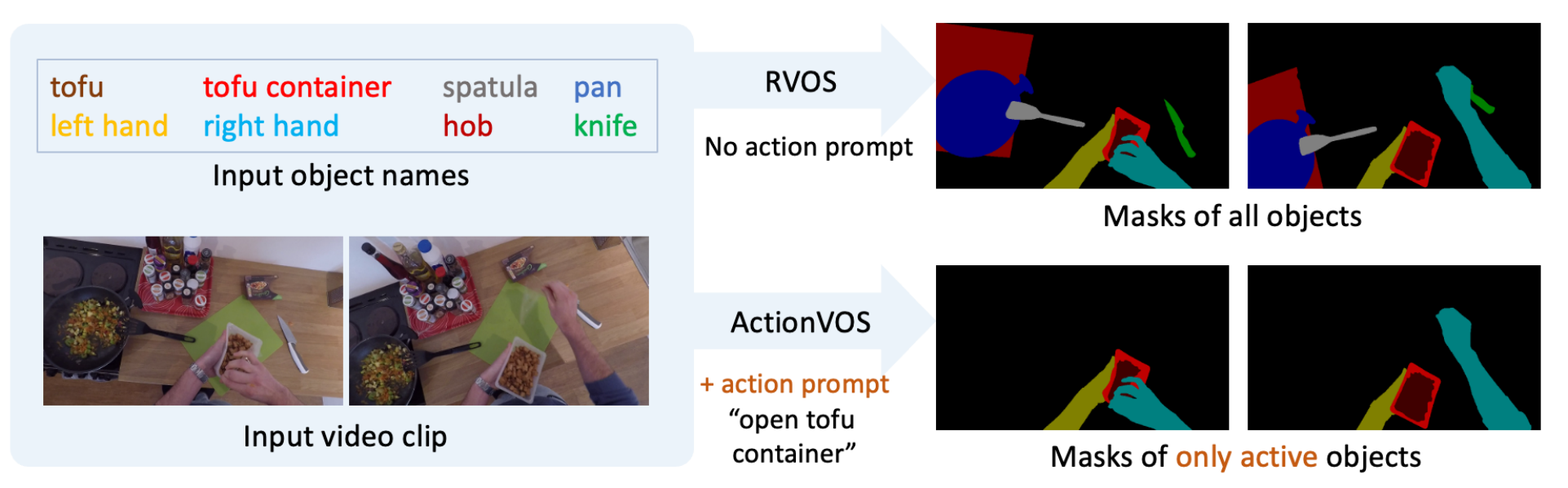}
  \caption{Comparison between ActionVOS and conventional RVOS settings.}
  \label{fig:input-output}
\end{figure}

Comparing with conventional RVOS tasks, ActionVOS focuses on if the referring object is interacted in the ongoing human action. We define the objects interacted in the action as positive $\mathcal{O_P}$, while other objects are negative $\mathcal{O_N}$. For the positive objects, all of them should be segmented through all frames. For the negative objects, their mask predictions should be all-zero since they do not participate in the action.
\cref{fig:input-output} compares the inputs and outputs of ActionVOS with conventional RVOS settings. Compared to RVOS, ActionVOS incorporates additional action prompts as input, expressed as an action narration ``open tofu container''. In this example, only active objects, \ie, hands, tofu, and tofu container are segmented in ActionVOS outputs.

\begin{figure}[t]
  \centering
  \includegraphics[width=\linewidth]{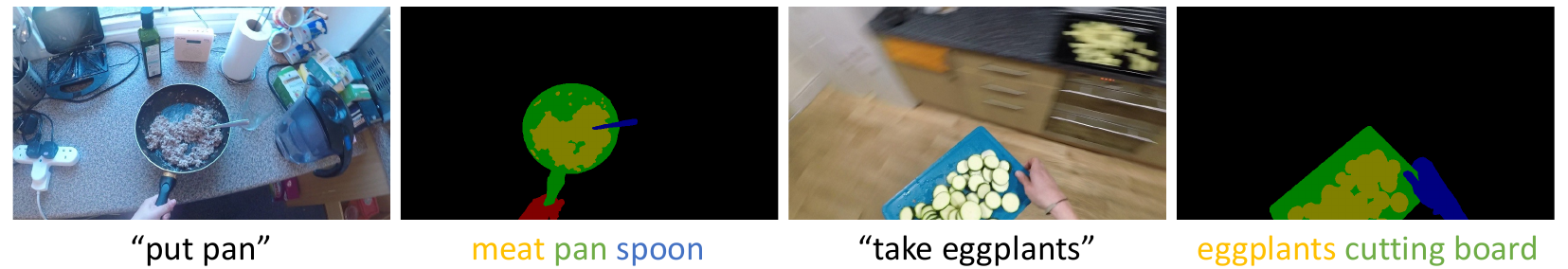}
  \caption{Examples of positive objects in ActionVOS.}
  \label{fig:positive-object}
  \vspace{-5mm}
\end{figure}

\noindent\textbf{Definition of ``positive''.} One of the most important concepts of ActionVOS is the definition of positive objects. According to the action prompt, we define positive objects as follows:
\begin{enumerate}[label={\arabic*)}]
\vspace{-2mm}
  \item Objects described by the action narration.
  \item Hands and hand-tools used for the action.
  \item Containers and contents interacted in the action.
\vspace{-2mm}
\end{enumerate}
1) The objects described by action narration are unquestionably defined as positive. 
2) Hand-tools, being operated by human during the action, are thus defined as positive. 
3) Taking the action ``put pan'' in \cref{fig:positive-object} as an example, we need to segment the objects inside the pan as target objects (\eg, meat, spoon). As the objects inside the pan are also put down with the ``pan'' through the action ``put'', they are subjected to the action of ``put pan''.  
Similarly, if an object moving with a container is mentioned in the action narration while the container itself is not, the container should also be subjected to the action. 
Therefore, we define containers and contents interacted in the action as positive. 
As shown in \cref{fig:positive-object}, the ``cutting board'' is defined as positive object for the action ``take eggplants''. On the other hand, if the ``cutting board'' as a moving vessel of ``eggplants'' is not defined as positive, this action becomes non-existent.    

\noindent\textbf{Data \& annotation.} Existing video object segmentation datasets such as VISOR \cite{darkhalil2022epic}, VOST \cite{tokmakov2023breaking} and VSCOS \cite{yu2023video} are collected on egocentric videos, providing both semantic segmentation labels and human action narrations \cite{damen2018scaling,damen2022rescaling,grauman2022ego4d}.
As VOST and VSCOS focus on objects undergoing state changes, they have only one active object being annotated for each action.
We only use the validation sets of these two datasets to evaluate ActionVOS performance on state-changed objects.
VISOR annotates a set of objects masks for each action, but lack precise indication of objects' involvement in actions, \ie, positive and negative classification labels.
To address this issue, we propose a labeling module in \cref{sec:labeling} to generates such classification labels with existing annotations.

\section{Proposed Method}
\label{sec:method}

\begin{figure}[tb]
  \centering
  \includegraphics[width=\linewidth]{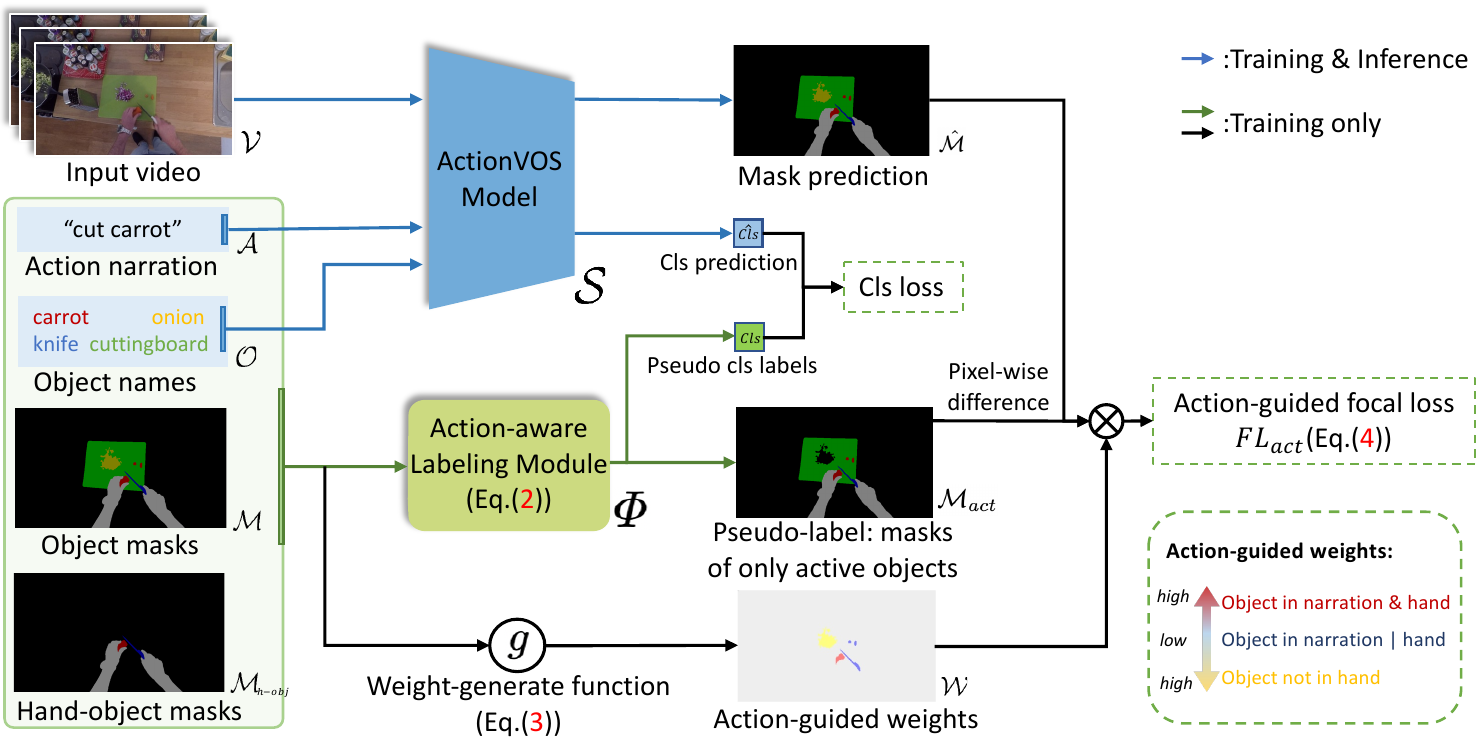}
  \caption{Overview of the proposed method.}
  \label{fig:proposed-method}
  \vspace{-5mm}
\end{figure}

As illustrated in \cref{fig:proposed-method}, we propose a method for the ActionVOS setting. In \cref{sec:model}, we develop an ActionVOS model $\mathcal{S}$, which is constructed by adding an extra classification head to an RVOS model. In \cref{sec:labeling}, we propose an action-aware labeling module $\Phi$. 
This module generates pseudo-labels of active objects, addressing the problem that \textit{existing datasets lack indication of objects' involvement in actions}. 
In \cref{sec:loss}, we propose an action-guided focal loss to reduce the impact of false positives from the generated pseudo-labels.

\noindent\textbf{Dataflow.}
During training, our method takes video $\mathcal{V}$, action narration $\mathcal{A}$, object names $\mathcal{O}$, object masks $\mathcal{M}$, and hand-object masks $\mathcal{M}_{h-obj}$ as input, which are all from existing annotations.  
An ActionVOS model $\mathcal{S}$ outputs classification of objects' involvement $\hat{Cls}\in[0,1]$ and mask predictions $\hat{\mathcal{M}}$, \ie, $\hat{Cls},\hat{\mathcal{M}}=\mathcal{S}(\mathcal{V},\mathcal{A},\mathcal{O})$. 
Given the input, the action-aware labeling module $\Phi$ generates pseudo-labels correspondingly, \ie, $Cls,\mathcal{M}_{act}=\Phi (\mathcal{A},\mathcal{O},\mathcal{M},\mathcal{M}_{h-obj})$. 
Along with the labeling module $\Phi$, a generating function $g$ generates pixel-wise weights for segmentation loss, \ie, $\mathcal{W} = g(\mathcal{A},\mathcal{O},\mathcal{M},\mathcal{M}_{h-obj})$.

\subsection{ActionVOS Model}
\label{sec:model}

We construct an ActionVOS model $\mathcal{S}$ by adding an extra classification head to an RVOS model.
Following state-of-the-art RVOS models \cite{wu2022language,yan2023universal}, which use a classification head to enhance segmentation performance, we add this classification head to distinguish positive and negative objects. This head predicts $\hat{Cls}(O_i)\in[0,1]$ indicating the probability of object $O_i$ being positive in action $\mathcal{A}$. During inference, we set a threshold $\theta$ to determine an object's positivity. Specifically, $\hat{Cls}(O_i)$ is used to adjust segmentation results $\mathcal{\hat{M}}(O_i)$ as follows:
\vspace{-2mm}
\begin{equation}
\vspace{-2mm}
    \small
    \label{eq:inference}
    \text{$\mathcal{\hat{M}}(O_i)$}=
    \begin{cases}
        \text{$\mathcal{\hat{M}}(O_i)$}, & \text{$\hat{Cls}(O_i)\geq \theta$} \\
        \text{0}, & \text{$\hat{Cls}(O_i)<\theta$},
    \end{cases}
\end{equation}

\noindent where $\theta$ is set as 0.75 according to experiments in \cref{exp:ablation}.

\subsection{Action-aware Labeling Module}
\label{sec:labeling}

To address the problem that \textit{existing datasets lack indication of objects' involvement in actions}, we propose an action-aware labeling module $\Phi$ to generate pseudo-labels of objects' involvement.
By using annotations of action narrations, semantic segmentation and hand-object segmentation, we label three types of objects as positive based on the guidance of action narrations and hand-object masks as follows:
\begin{enumerate}[label={\arabic*)}]
\vspace{-2mm}
  \item Objects mentioned in the action narrations.
  \item Objects inside hand-object masks.
  \item Objects that intersect with hand-object bounding boxes.
\vspace{-2mm}
\end{enumerate}
\noindent These three types correspond to the three definitions of ``positive'' in \cref{sec:task}.
For type 3), such a design identifies a large number of objects that are potentially positive, because these objects within close reach are highly likely to be relevant to the action, such as containers and contents.

Pseudo-labels generated by the labeling module $\Phi$ contains two parts: classification labels $Cls$ and action-aware object masks $\mathcal{M}_{act}$. For each object $O_i$, its pseudo-label is formulated as:

\vspace{-5mm}
\begin{equation}
\vspace{-2mm}
\label{eq:labeling}
\small
\begin{aligned}
    &Cls(O_i)= \begin{cases}
        1, & O_i \in \mathcal{A} \\
        1, & \mathcal{M}(O_i) \in  M_{h-obj} \\
        1, & \mathcal{M}(O_i) \cap B_{h-obj} \neq \emptyset \\
        0, & \text{otherwise},
    \end{cases}
    &&\mathcal{M}_{act}(O_i)= \begin{cases}
        \mathcal{M}(O_i), & Cls(O_i)=1 \\
        0, &Cls(O_i)=0,
    \end{cases}
\end{aligned}
\end{equation}

\noindent where $B_{h-obj}$ stands for the minimal bounding box of hand-object mask $\mathcal{M}_{h-obj}$.
The generated pseudo-labels $Cls,\mathcal{M}_{act}$ are used to train the ActionVOS model.  

However, since \cref{eq:labeling} provides a more relaxed definition of positive compared to \cref{sec:task}, practical issues may arise. 
While it correctly identifies many potential positives, it also introduces multiple false positives simultaneously. To address this issue, we introduce an action-guided focal loss in \cref{sec:loss} aimed at reducing the impact of false positives.

\subsection{Action-guided Focal Loss}
\label{sec:loss}
To reduce the impact of false positives, action-guided focal loss $FL_{act}$ is proposed by adding pixel-wise action-guided weights $\mathcal{W}$ to segmentation focal loss $FL$ \cite{lin2017focal}.

\begin{figure}[tb]
  \centering
  \includegraphics[width=\linewidth]{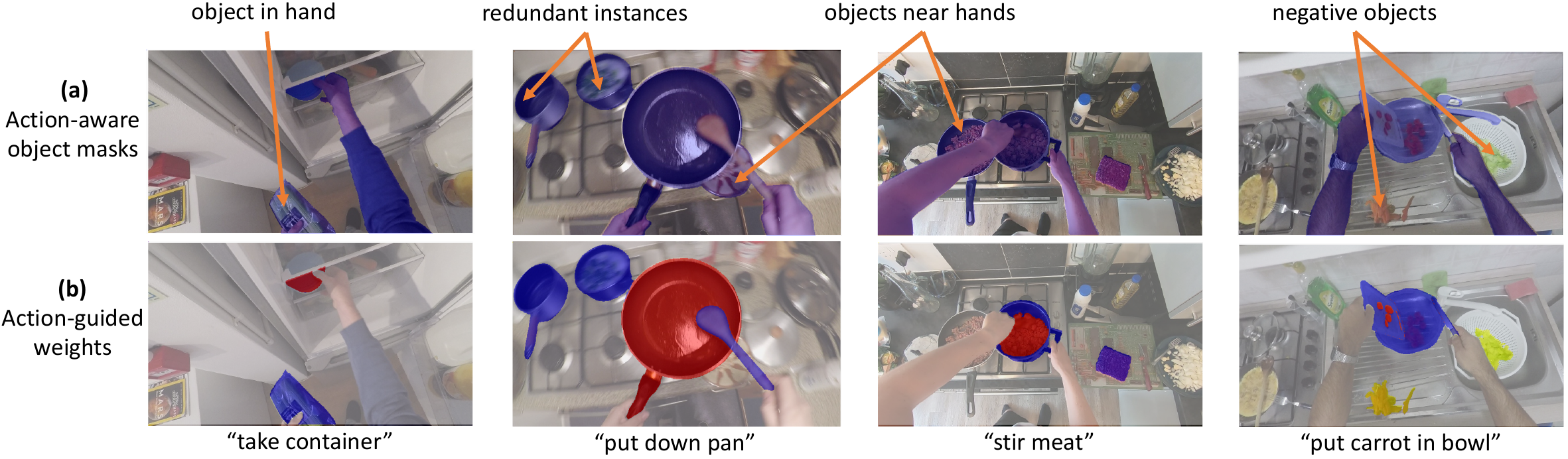}
  \caption{Action-aware object masks and action-guided weights. In action-guided weights, $\lambda_{pos}$ is in red, $\lambda_{nar}$ and $\lambda_{h-obj}$ are in blue, $\lambda_{neg}$ is in yellow. }
  \label{fig:proposed-loss}
  \vspace{-5mm}
\end{figure}

\cref{fig:proposed-loss} (a) analyzes typical mistakes in action-aware object masks generated from $\Phi$.  
In the ``take container'' example, although the object is in contact with left hand, it is not involved in this action. In the ``put down pan'' example, redundant instances of ``pan'' are not active even though they are mentioned by the narration.

To address these false positives, we adjust the weight when calculating the pixel-level segmentation loss, to make the segmentation model aware of interacted positive objects.
We establish three rules: 
\begin{enumerate}[label={\arabic*)}]
\vspace{-2mm}
\item Objects in both action narration and hand-object bounding box > those solely in either.
\item Objects mentioned in action narration or in contact with hands > those only intersecting with hand-object bounding boxes. 
\item Objects labeled as negative are assigned a high weight within their masks.
\vspace{-2mm}
\end{enumerate}
The rules 1,2 aim to prioritize objects more likely to be actively involved among all potential positives, while the last rule penalizes the model for mis-segment negative objects as positive.
As shown in \cref{fig:proposed-loss} (b), the ``container'' (1st column), both mentioned in narration and in contact with right hand, receives higher weights than the left-hand object, which is only in contact with left hand (according to rule 1).
The ``pot'' (3rd column), held by right hand, receives higher weight than the ``pan'' under left hand (according to rule 2). 

Following these rules, we propose a generating function of action-guided weights $\mathcal{W}$. For a pixel $(h,w)$ inside object $O_i$'s region, \ie, $M_t(O_i,h,w) = 1$, its action-guided weight is: 
\vspace{-3mm}
\begin{equation}
\vspace{-2mm}
    \small
    W_t(O_i,h,w) = 
    \begin{cases}
         \text{$\lambda_{pos},$} & \text{$O_i\in\mathcal{A}$ and $B_{h-obj,t}(h,w) = 1$} \\
        \text{$\lambda_{nar}$}, & \text{$O_i\in\mathcal{A}$ and $B_{h-obj,t}(h,w) = 0$} \\
        \text{$\lambda_{h-obj}$}, & \text{$O_i\notin\mathcal{A}$ and $M_{h-obj,t}(h,w) = 1$} \\
        \text{$\lambda_{neg}$}, & \text{$O_i\notin\mathcal{A}$ and $B_{h-obj,t}(h,w) = 0$} \\
        \text{1}, & \text{otherwise.}
    \end{cases}
    \label{eq:weight}
\end{equation}
\noindent The relationship of these weights are defined as $\lambda > 1$, $\lambda_{pos} > \lambda_{nar}$, and $\lambda_{pos} > \lambda_{h-obj}$. Empirically, we set $\lambda_{pos}=5$, $\lambda_{nar}=2$, $\lambda_{h-obj}=2$, $\lambda_{neg}=5$. For pixels outside $O_i$'s region, where $M_t(O_i,h,w) = 0$, the weight is set as $1$.

The action-guided weights are added to the pixel-level segmentation focal loss \cite{lin2017focal}. Let $p_{h,w}\in[0,1]$ be the probability for pixel $(h,w)$'s positivity, which is predicted from the model's output $\mathcal{\hat{M}}$. $y_{h,w}\in\{0,1\}$ is the generated pseudo-labels of the pixel on the same location, coming from $\mathcal{M}_{act}$. The action-guided focal loss $FL_{act}$ for each frame is expressed as:
\vspace{-2mm}
\begin{equation}
\vspace{-2mm}
\small
\begin{aligned}
    p_t & = p_{h,w}\cdot y_{h,w} + (1-p_{h,w})\cdot(1-y_{h,w}), \\
    \alpha_t & = \alpha\cdot y_{h,w} + (1-\alpha)\cdot(1-y_{h,w}), \\
    FL_{act} & = -\frac{1}{H*W} \sum_{h=1,w=1}^{H,W} W_t(O_i,h,w) \cdot \alpha_t \cdot (1 - p_t)^\gamma \cdot \log(p_t),
\end{aligned}\label{eq:act-loss}
\end{equation}
\noindent where $\alpha$ is a balanced parameter, $\gamma$ is a focusing parameter. Following previous work \cite{wu2022language}, we set $\alpha=0.25$ and $\gamma=2$. 

As \cref{eq:labeling,eq:act-loss} indicates, both the action-guided focal loss and action-aware labeling module has no trainable parameters. 
This parameter-free design allows our method to segment active objects with existing readily-available data. In addition, as our method's inputs and outputs align with the conventional RVOS task, it is compatible with existing RVOS models.

Hand-object masks $\mathcal{M}_{h-obj}$ used in \cref{eq:labeling,eq:weight} are obtained from the human annotation in VISOR \cite{darkhalil2022epic} dataset. Hand-object masks are only used in training for generating pseudo-labels. During inference, ActionVOS model does not take hand-object masks as input and has no need to estimate the contact with hands. 

\section{Experiments}
\label{sec:experiments}

\subsection{Datasets}

\noindent{\bf VISOR \cite{darkhalil2022epic}} is a new dataset conducted on EPIC-KITCHENS \cite{damen2018scaling,damen2022rescaling} suitable for segmenting hands and active objects in egocentric videos. We use their videos and annotations for both training and validation. We exclude videos annotated with less than 2 frames. In the validation set, we randomly choose 330 action clips and manually annotate the positive and negative objects. 

\noindent{\bf VOST \cite{tokmakov2023breaking}} is a recent dataset collected for video object segmentation under transformations. We only use VOST for validation since only one object class is annotated for each video. VOST annotate multiple instances and we treat all instances within the same video as one active object.

\noindent{\bf VSCOS \cite{yu2023video}} is constructed recently by selecting state-changing videos from EPIC-KITCHENS \cite{damen2018scaling,damen2022rescaling}. We also only use VSCOS for validation of state-changed objects. As it shares multiple video clips with VISOR, we filter out the video clips who have appeared in the training set of VISOR to avoid data leakage.

For all three datasets, we adhere to their original split rules for dividing the train-valid set. After the pre-processing, we obtain 13,205 videos and 76,873 objects for training, 467 videos and 1,841 objects for validation. The validation sets contain 1,133 positive and 708 negative objects.

\subsection{Evaluation Metrics}
Following \cite{liu2023gres}, we employ mean IoU (mIoU), cumulative IoU (cIoU), generalized IoU (gIoU) and a classification accuracy (Acc) as evaluation metrics.

\noindent{\bf mIoU and cIoU.}
mIoU and cIoU are widely-used in segmentation tasks \cite{lin2014microsoft,mottaghi2014role,zhou2017scene,wu2020phrasecut,yang2022lavt,liu2023gres}. mIoU calculates the mean intersection over union while cIoU calculates the total intersection pixels over total union pixels. 
As ActionVOS introduces a novel concept of distinguishing positive and negative objects, we report mIoU and cIoU separately for positive and negative objects, \ie, p-mIoU, n-mIoU, p-cIoU, n-cIoU.

\noindent{\bf gIoU.}
gIoU is introduced in \cite{liu2023gres} to combine the segmentation result and a no-target classification result. In our work, this metric simultaneously evaluates the ability to segment positive objects and distinguish negative objects. 

\noindent{\bf Acc.}
We further use a classification accuracy to evaluate the model's performance on identifying active objects. It is calculated by binary classification results, Acc$=\frac{TN+TP}{TN+TP+FN+FP}$.

\subsection{Implementation Details}
{\bf Model Settings.}
We apply ReferFormer \cite{wu2022language} with different visual backbones as baseline RVOS models in our experiments. The backbone of ReferFormer can be replaced with ResNet-101 \cite{he2016deep}, Swin-L \cite{liu2021swin}, or Video-Swin-Base \cite{liu2022video}. RoBERTa \cite{liu2019roberta} is employed as the text encoder, where its parameters are re-trained in our experiment. The extra classification head is a linear layer, which receives averaged features from the last output layer \cite{carion2020end,zhu2020deformable} to predict binary classification, defined by \textit{nn.Linear(256,1)} in pytorch \cite{paszke2019pytorch} implementation.

\noindent{\bf Training Details.}
All models are trained from best checkpoints on Refer-YouTube-VOS \cite{seo2020urvos} benchmarks. We follow all the training settings of ReferFormer \cite{wu2022language}, including epochs, optimizer \cite{loshchilov2018decoupled}, loss coefficients \cite{rezatofighi2019generalized,milletari2016v,lin2017focal} and data augmentations \cite{wang2021end}. We replace the segmentation focal loss \cite{lin2017focal} with our proposed action-guided focal loss, and introduce a binary cross-entropy loss to train the additional classification head.  
The weight for the extra classification loss is set to $2$.

\subsection{ActionVOS Results}

\noindent{\bf Quantitative results on VISOR.} We analyze the segmentation performance of ActionVOS models. Here, the results of RVOS is provided as the upper bound of p-mIoU/p-cIoU, as it treats all objects as positive. As shown in \cref{tab:actionvos-results}, compared to RVOS, \proposed offers significant n-mIoU/n-cIoU decrease while ensuring there is only a slight decrease in p-mIoU/p-cIoU. This indicates that ActionVOS significantly reduces the mis-segmentation of non-interacted objects while keeping the ability of segmenting active objects. 
We also evaluate our method by removing action prompts input and training with only object names. For example, we replace the language input ``knife used in the action of cut apple'' with ``knife'', while keeping the same pseudo-labels and loss weights. Experimental results show that the models trained with action prompts achieve much better performance, confirming that action prompts help \proposed models focus on active objects. The ActionVOS models trained without action prompts perform much lower p-IoUs than RVOS models. This is because the training of these models is misled by the negative pseudo-labels (all-zero for negative objects). For example, a ``knife'' is positive in the action ``cut apple'', but negative in ``open fridge''. Without action prompts, ActionVOS model has no evidence to distinguish knives' positivity. In contrast, RVOS treats all objects as positive, avoiding the impact of negative pseudo-labels.

\cref{tab:actionvos-results} also shows the results of replacing the backbone network with different structures, including ResNet-101 \cite{he2016deep}, Swin-L \cite{liu2021swin}, and Video-Swin-Base \cite{liu2022video}.  
With our proposed setting, all backbones achieve significant improvements on these metrics. Quantitative results demonstrate that the proposed \proposed is compatible with various existing network structures.

\begin{table*}[tb]
\caption{Quantitative results of \proposed on VISOR. ``AP" indicates whether action prompts are used for training. ``RF" stands for ReferFormer. * indicates RVOS is the upper bound of pos-mIoU under this experimental setting.}
\vspace{-3mm}
\small
  \centering
  \setlength{\tabcolsep}{1mm}{
  \begin{tabular}{@{}llccccccc@{}}
    \toprule
    Model & Setting & AP & p-mIoU $\uparrow$ & n-mIoU $\downarrow$ & p-cIoU $\uparrow$ & n-cIoU $\downarrow$ & gIoU $\uparrow$ & Acc $\uparrow$\\
    \midrule
     & RVOS* &   & 67.7 & 54.2 & 73.2 & 67.5 & 43.8 & 59.1\\
     \cline{2-9} \specialrule{0em}{.5pt}{.5pt}
    RF-R101 & \proposed &  \xmark & 56.3 & 19.9 & 61.2 & 32.8 & 66.8 & 72.9\\
     & \proposed & \cmark & \textbf{65.4} & \textbf{19.0} & \textbf{72.4} & \textbf{32.7} & \textbf{70.9} & \textbf{82.4}\\
    \midrule
     & RVOS* &   & 71.8 & 59.7 & 79.9 & 73.0 & 46.8 & 59.4\\
     \cline{2-9} \specialrule{0em}{.5pt}{.5pt}
    RF-SwinL & \proposed & \xmark  & 64.4 & 28.2 & 71.8 & 47.3 & 65.1 & 72.8\\
     & \proposed & \cmark & \textbf{69.1} & \textbf{24.6} & \textbf{75.7} & \textbf{46.5} & \textbf{70.3} & \textbf{80.7}\\  
    \midrule
     & RVOS* &   & 70.5 & 58.5 & 78.1 & 71.7 & 45.6 & 59.2\\
     \cline{2-9} \specialrule{0em}{.5pt}{.5pt}
    RF-VSwinB & \proposed &  \xmark & 61.6 & 25.2 & 66.8 & 44.5 & 65.7 & 72.5\\
     & \proposed & \cmark & \textbf{68.2} & \textbf{22.0} & \textbf{75.0} & \textbf{41.5} & \textbf{70.6} & \textbf{81.2}\\
    \bottomrule
  \end{tabular}
  \label{tab:actionvos-results}}
  \vspace{-2mm}
\end{table*}

\noindent{\bf Quantitative results under object state changes.} We evaluate \proposed on the VOST and VSCOS datasets, which consist of object state changes. As demonstrated in \cref{tab:vost-vscos-results}, the ActionVOS model outperforms RVOS in segmentation performance across both datasets. This suggests that our method effectively handles scenarios involving object state changes, with action prompts providing enhanced understanding of state changing.

\begin{table*}[tb]
\caption{Comparison with RVOS model under scenarios with object state changes.}
\vspace{-3mm}
\small
\centering
\setlength{\tabcolsep}{1.5mm}{
\begin{tabular}{@{}lccccc@{}}
\toprule
\multirow{2}{*}{Setting} & \multirow{2}{*}{AP} & \multicolumn{2}{c}{\underline{VOST}} & \multicolumn{2}{c}{\underline{VSCOS}} \\
& & p-mIoU $\uparrow$ & p-cIoU $\uparrow$ & p-mIoU $\uparrow$ & p-cIoU $\uparrow$\\
\midrule
 RVOS &  \xmark & 29.3 & 17.5 & 46.4 & 44.9\\
\proposed &  \xmark & 9.0 & 8.2 & 22.5 & 33.0\\
\proposed & \cmark & \textbf{32.3} & \textbf{22.8} & \textbf{49.4} & \textbf{49.6}\\
\bottomrule
\end{tabular}
\label{tab:vost-vscos-results}}
\vspace{-7mm}
\end{table*}

\noindent\textbf{Comparison with baseline methods.} We compare ActionVOS model with baseline methods on three datasets. The baseline methods are: \textbf{1)Hand-object segmentation (HOS) model.} We take the best HOS model in \cite{darkhalil2022epic} as a baseline model, which is also trained on VISOR dataset. HOS model segments hands and hand-objects, and we treat the segmentation results as positive object masks.  \textbf{2)RVOS+S4.2.} We use \cref{eq:labeling} in \cref{sec:labeling} as a post-process of RVOS model outputs. Note that \cref{eq:labeling} take ground-truth hand-object masks as input. The comparisons are shown in \cref{tab:baselines}. ActionVOS model outperforms other baselines in terms of positive IoUs, gIoU and accuracy. HOS model has lower negative IoUs, because it only segment hands and hand-objects. RVOS+S4.2 shows worse results since S4.2 brings amounts of false positives.

\begin{table}[h]
\caption{Comparion with HOS \cite{darkhalil2022epic} and RVOS\cite{wu2022language}+S4.2. RVOS model is RF-R101.}
\scriptsize
  \centering
  \begin{tabular}{@{}lcccccccccc@{}}
    \toprule
    \multirow{2}{*}{Method} & \multirow{2}{*}{p-mIoU $\uparrow$} & \multirow{2}{*}{n-mIoU $\downarrow$} & \multirow{2}{*}{p-cIoU $\uparrow$} & \multirow{2}{*}{n-cIoU $\downarrow$} & \multirow{2}{*}{gIoU $\uparrow$} & \multirow{2}{*}{Acc $\uparrow$} & \multicolumn{2}{c}{\underline{VOST}} & \multicolumn{2}{c}{\underline{VSCOS}}\\
    & & & & & & & p-mIoU & p-cIoU & p-mIoU & p-cIoU\\
    \midrule
    \text{HOS} & 56.2 & \textbf{11.4} & 58.1 & \textbf{16.8} & 68.8 & 77.0 & 19.4 & 13.1 & 34.4 & 24.1\\
    RVOS+S4.2 & 65.3 & 35.2 & 71.5 & 56.4 & 60.4 & 75.1 & 29.3 & 17.5 & 46.4 & 44.9\\
    ActionVOS & \textbf{65.4} & 19.0 & \textbf{72.4} & 32.7 & \textbf{70.9} & \textbf{82.4} & \textbf{32.3} & \textbf{22.8} & \textbf{49.4} & \textbf{49.6}\\
    \bottomrule
  \end{tabular}
  \label{tab:baselines}
  \vspace{-2.5mm}
\end{table}

\begin{figure}[h]
  \centering
  \includegraphics[width=\linewidth]{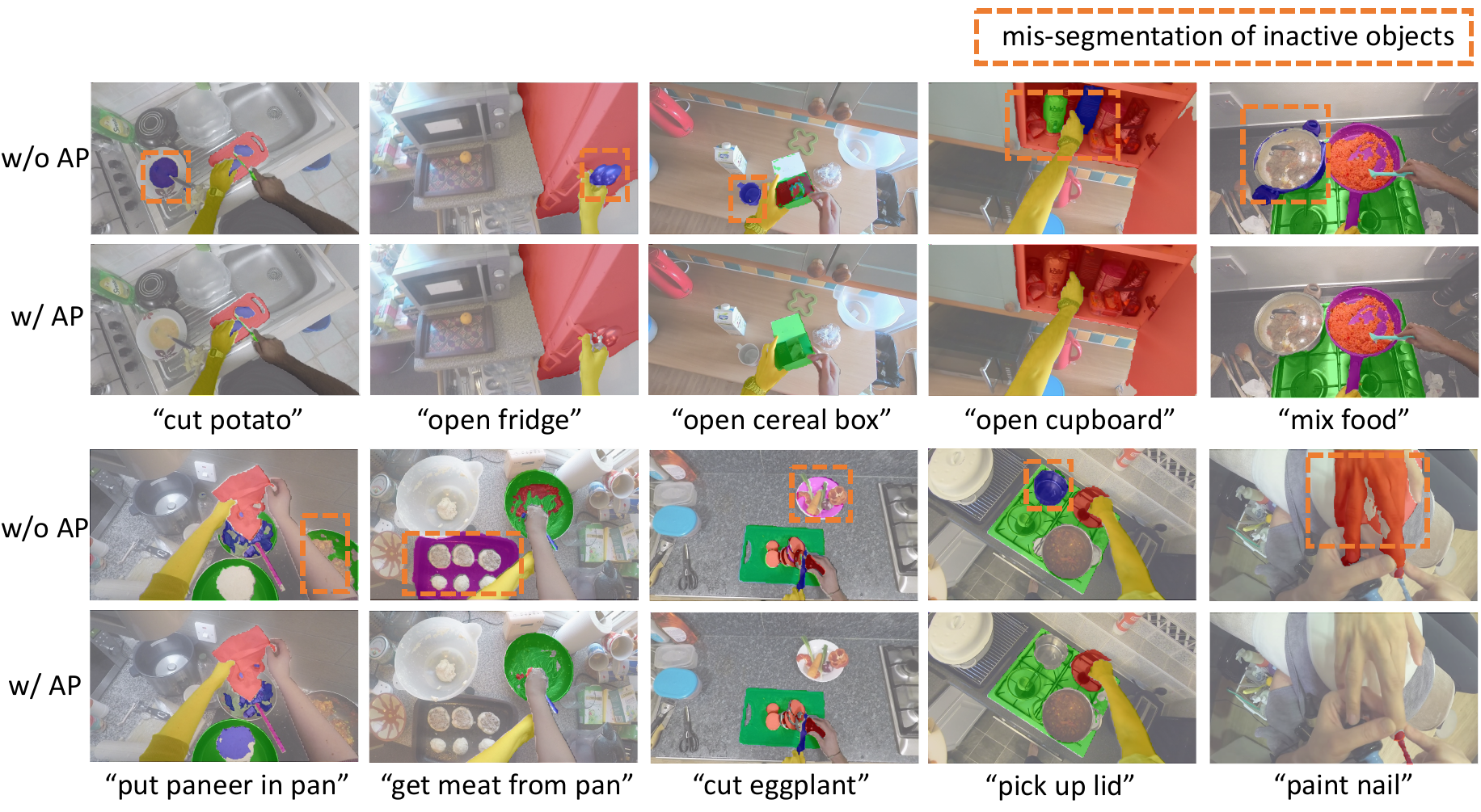}
  \caption{Visualization results of ActionVOS models trained w/ and w/o action prompts.}
  \label{fig:result-action}
  \vspace{-3mm}
\end{figure}

\subsection{Qualitative results} 
\textbf{Action prompts.} \cref{fig:result-action} compares segmentation predictions from models trained with and without action prompts. Given the same input object names, model trained with action prompts correctly identifies objects involved in the action. In contrast, model trained with only object names tend to segment inactive objects, \eg, redundant instances.

\begin{figure}[tb]
  \centering
  \includegraphics[width=\linewidth]{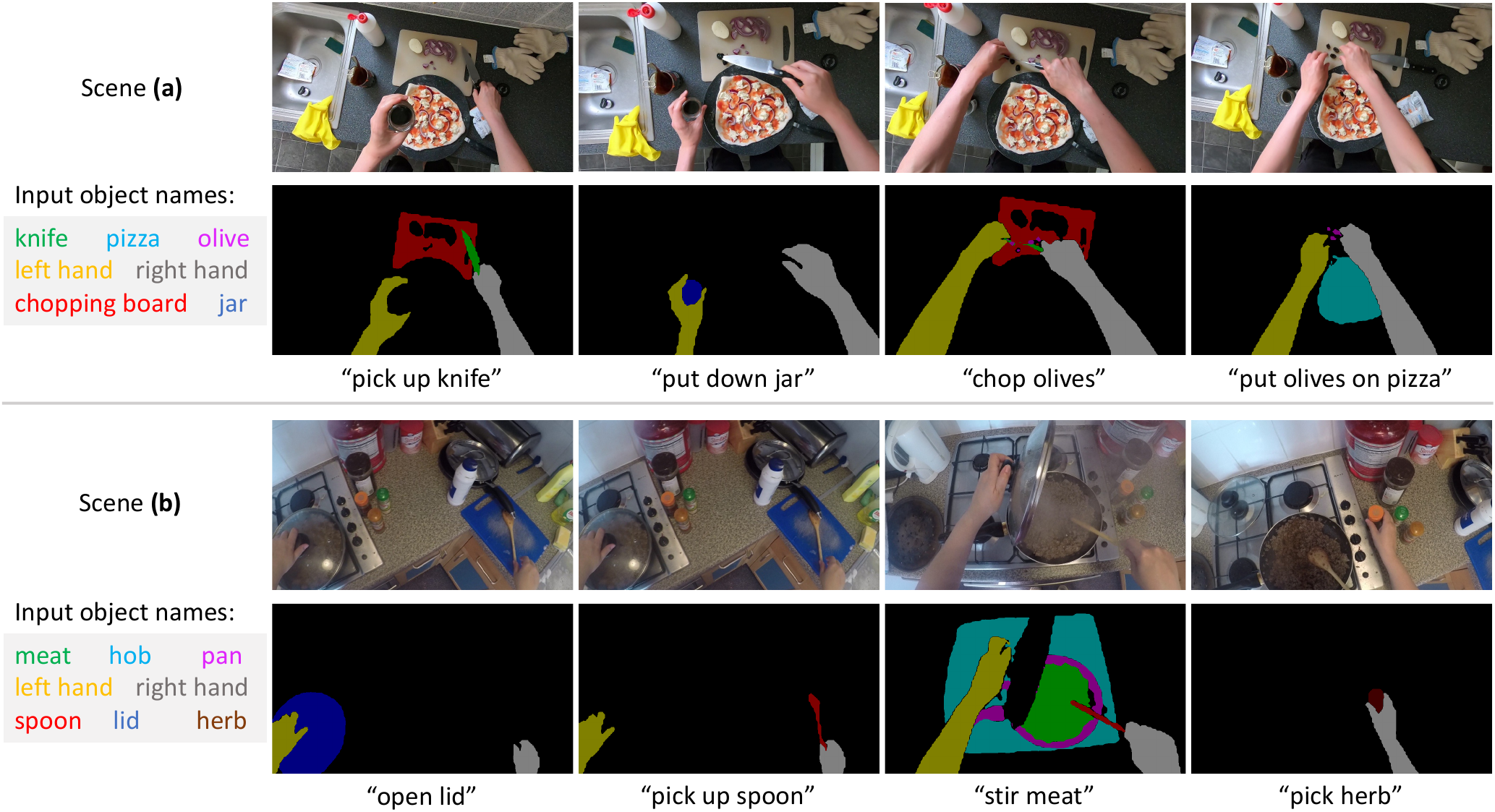}
  \caption{Segmentation results of \proposed in the same scenes. For each video clip, all frames share the same input object names.}
  \label{fig:result-timeline}
  \vspace{0mm}
\end{figure}

\noindent\textbf{Effect of action prompts in identical scenes.} \cref{fig:result-timeline} shows \proposed's segmentation results in the same scenes. Even though those actions occur within the same scene and share identical input object names, our method still correctly segments active objects. This underscores the model's comprehension of human-object interaction, facilitated by action narrations.

\begin{figure}[tb]
\begin{minipage}{0.59\linewidth}
\centering
\includegraphics[width=\linewidth]{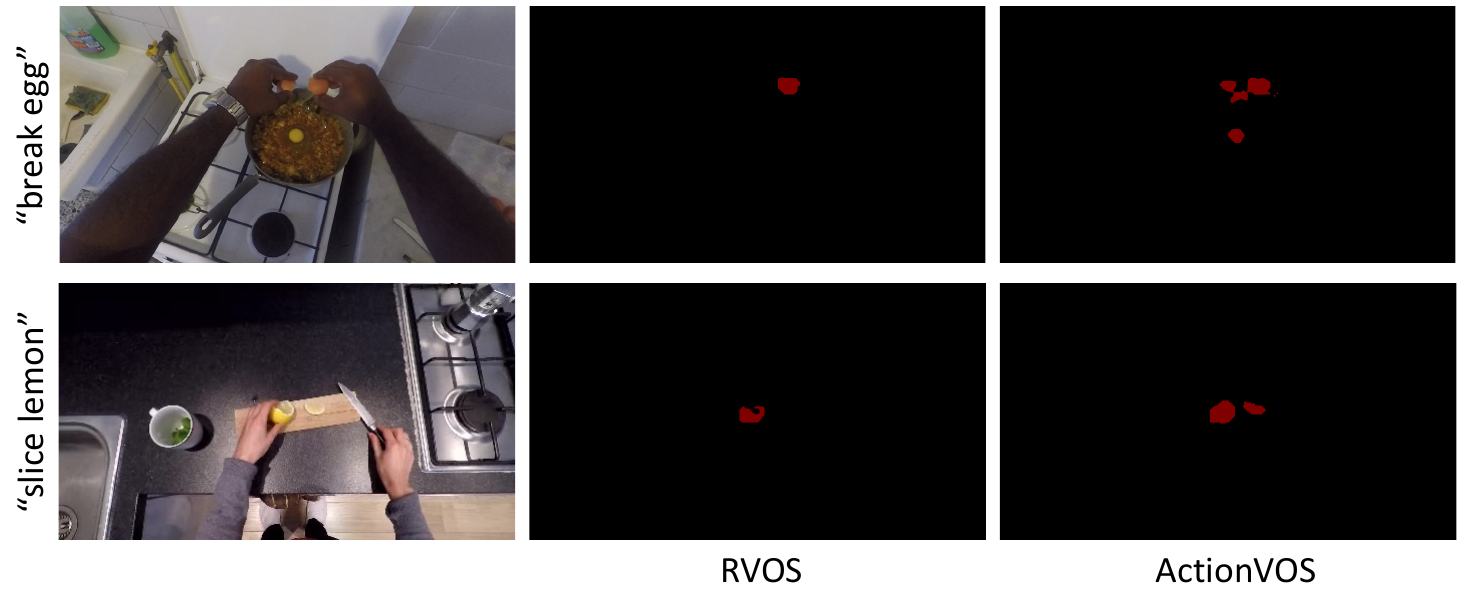}
\caption{Segmentation results of ActionVOS under scenarios with object state changes.}
\label{fig:result-state}
\end{minipage}\hfill
\begin{minipage}{0.4\linewidth}
\centering
\includegraphics[width=\linewidth]{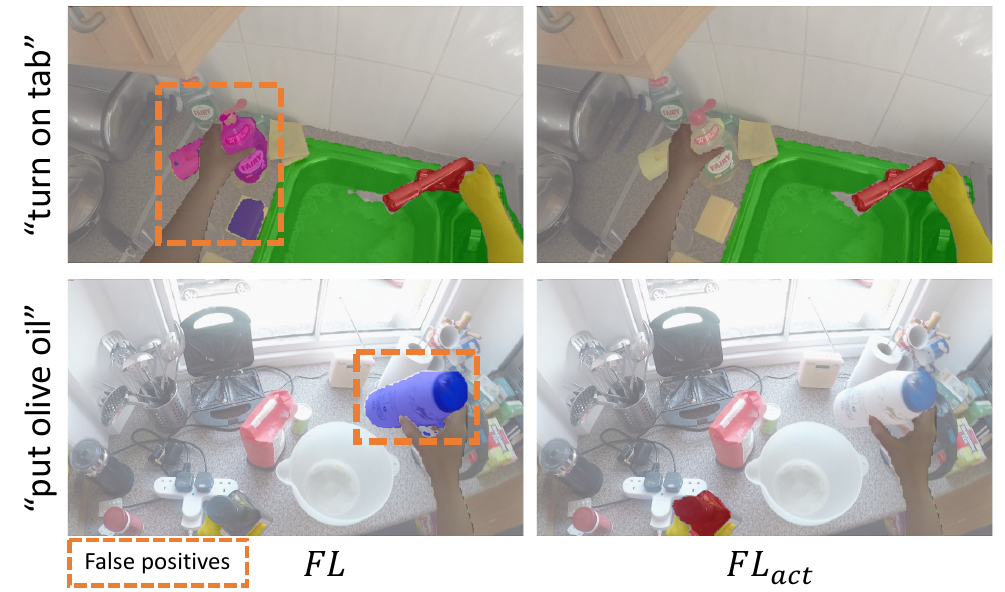}
\caption{The effect of the proposed action-guided focal loss $FL_{act}$.}
\label{fig:result-loss}
\end{minipage}\hfill
\end{figure}

\noindent\textbf{Effect of action prompts under state changes.} \cref{fig:result-state} visualizes the segmentation results of ActionVOS in comparison with RVOS under scenarios with object state change.
RVOS method fails to segment objects after a change of state, such as broken eggshells and yolks, and sliced lemons. In contrast, our method successfully identifies these state-changed objects, confirming that action prompts help to enhance understanding of state changing.

\subsection{Action Vocabulary}

\noindent\textbf{Vocabulary statistics} of training \cite{darkhalil2022epic} and validation \cite{darkhalil2022epic,tokmakov2023breaking,yu2023video} sets with the number of unseen categories are provided in \cref{tab:action-classes}.

\begin{table}[h]
\vspace{-6mm}
\caption{Vocabulary statistics.}
\vspace{-2mm}
\small
  \centering
  \setlength{\tabcolsep}{3.5mm}
  \begin{tabular}{@{}lcccccc@{}}
    \toprule
     Split& \# actions & \# verbs & \# nouns\\
    \midrule
    Training \cite{darkhalil2022epic} & 1898 & 90 & 242\\
    Validation \cite{darkhalil2022epic,tokmakov2023breaking,yu2023video} & 187 / 39 / 37 & 43 / 20 / 9 & 125 / 32 / 33 \\
    Unseen in validation & 37 / 26 / 9 & 0 / 3 / 0 & 4 / 16 / 2 \\
    \bottomrule
  \end{tabular}
  \label{tab:action-classes}
  \vspace{-5mm}
\end{table}

\noindent\textbf{Evaluation on unseen categories.} We compare ActionVOS model performance with other baselines on unseen actions in \cref{tab:unseen-results}, where our method achieved best results.
This is because the ActionVOS model not only identifies target objects through input object names, but also learn to segment active objects through human action interactions. For example, in the last visualization in \cref{fig:result-action}, neither ``paint'' nor ``nail'' appear in the training set, while ActionVOS with action prompts still successfully segmented the painted nail.

\begin{table}[h]
\vspace{-5mm}
\caption{Evaluation on unseen actions.}
\vspace{-2mm}
\scriptsize
  \centering
  \begin{tabular}{@{}lcccccccccc@{}}
    \toprule
    \multirow{2}{*}{Method} & \multirow{2}{*}{p-mIoU $\uparrow$} & \multirow{2}{*}{n-mIoU $\downarrow$} & \multirow{2}{*}{p-cIoU $\uparrow$} & \multirow{2}{*}{n-cIoU $\downarrow$} & \multirow{2}{*}{gIoU $\uparrow$} & \multirow{2}{*}{Acc $\uparrow$} & \multicolumn{2}{c}{\underline{VOST}} & \multicolumn{2}{c}{\underline{VSCOS}}\\
    & & & & & & & p-mIoU & p-cIoU & p-mIoU & p-cIoU\\
    \midrule
    RVOS \cite{wu2022language}& 60.0 & 49.0 & 63.5 & 63.6 & 42.9 & 65.3 & 18.6 & 12.6 & 31.5 & 21.4\\
    HOS \cite{darkhalil2022epic} & 51.9 & \textbf{9.0} & 57.3 & \textbf{6.4} & 64.9 & 72.0 & 13.6 & 11.4 & 42.7 & 38.8\\
    ActionVOS & \textbf{60.3} & 21.0 & \textbf{65.7} & 39.7 & \textbf{66.1} & \textbf{79.7} & \textbf{22.5} & \textbf{18.0} & \textbf{44.9} & \textbf{43.1}\\
    \bottomrule
  \end{tabular}
  \label{tab:unseen-results}
  \vspace{-5mm}
\end{table}

\begin{figure}[h]
  \centering
  \includegraphics[width=\linewidth]{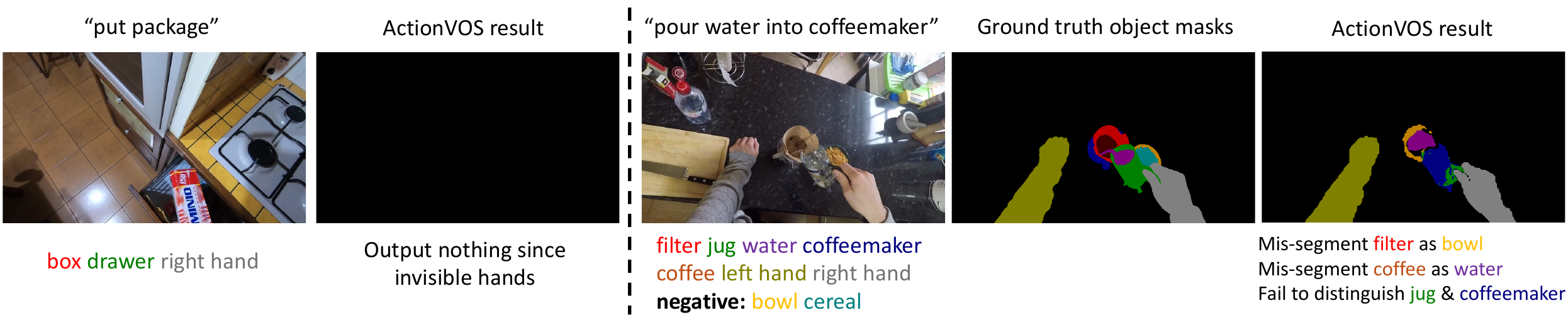}
  \caption{Visualization of ActionVOS failed cases.}
  \label{fig:result-fail}
\end{figure}

\noindent
\textbf{Hard action categories}. The actions with segmentation p-mIoU lower than 30\% in VISOR validation set are listed below: ``put down pakage'', ``dry hand'', ``put tea towel'', ``push oven tray'', ``pour-into water'', ``sprinkle-on salt'', ``take-out grape'', ``take carrot bag'', ``pick-up spinach'', ``get meat mix''. In VOST validation set, ``cut paper'' and ``divide dough'' get lowest p-mIoU. We find that \textbf{invisible hands, ambiguous object names and significant shape change} bring low ActionVOS results. The visualization of typical fail cases are shown in \cref{fig:result-fail}.   

\begin{table}[h]
\caption{Ablation study for the classification head and its threshold $\theta$.}
\vspace{-3mm}
\small
  \centering
  \setlength{\tabcolsep}{1.5mm}{
  \begin{tabular}{@{}lcccccc@{}}
    \toprule
     & p-mIoU $\uparrow$ & n-mIoU $\downarrow$ & p-cIoU $\uparrow$ & n-cIoU $\downarrow$ & gIoU $\uparrow$ & Acc $\uparrow$\\
    \midrule
     \textit{No head} & 67.4 & 31.2 & 73.4 & 57.5 & 63.2 & 76.0\\
    \midrule
    $\theta=0.25$ & \textbf{67.6} & 32.7 & \textbf{74.1} & 57.6 & 61.8 & 75.3\\
    $\theta=0.50$ & 67.5 & 30.6 & 74.1 & 55.9 & 63.5 & 76.7\\
    $\theta=0.75$ & 65.4 & \textbf{19.0} & 72.4 & \textbf{32.7} & \textbf{70.9} & \textbf{82.4}\\
    \bottomrule
  \end{tabular}
  \label{tab:ablation-classification}}
  \vspace{-2mm}
\end{table}

\begin{table}[h]
\caption{The impact of language prompts and fine-tuning text encoders.}
\vspace{-2.5mm}
  \centering
  \setlength{\tabcolsep}{1mm}{
  \begin{tabular}{@{}lccccccc@{}}
    \toprule
    Prompt & Tuned & p-mIoU $\uparrow$ & n-mIoU $\downarrow$ & p-cIoU $\uparrow$ & n-cIoU $\downarrow$ & gIoU $\uparrow$ & Acc $\uparrow$\\
    \midrule
    \textit{NoAction} & \xmark & 54.5 & 17.7 & 59.5 & 31.6 & 66.1 & 71.3\\
    \textit{+,Action} & \xmark  & 60.4 & 17.0 & 67.1 & \textbf{23.7} & 70.0 & 79.5\\
    \textit{+sAction} &  \xmark & \textbf{61.7} & \textbf{16.3} & \textbf{67.5} & 27.6 & \textbf{70.6} & \textbf{80.2}\\
    \midrule
    \textit{NoAction} & \cmark & 56.3 & 19.9 & 61.2 & 32.8 & 66.8 & 72.9\\
    \textit{+,Action} & \cmark & 65.1 & 19.6 & 70.9 & 34.5 & 70.8 & 82.4\\
    \textit{+sAction} & \cmark & \textbf{65.4} & \textbf{19.0} & \textbf{72.4} & \textbf{32.7} & \textbf{70.9} & \textbf{82.4}\\
    \bottomrule
  \end{tabular}
  \label{tab:ablation-prompt}}
  \vspace{-2mm}
\end{table}

\begin{table}[h]
\caption{The effect of proposed action-guided focal loss.}
\vspace{-2.5mm}
\small
  \centering
  \setlength{\tabcolsep}{1mm}{
  \begin{tabular}{@{}lcccccc@{}}
    \toprule
     & p-mIoU $\uparrow$ & n-mIoU $\downarrow$ & p-cIoU $\uparrow$ & n-cIoU $\downarrow$ & gIoU $\uparrow$ & Acc $\uparrow$\\
    \midrule
     $FL$ \cite{lin2017focal} & 65.4 & 20.5 & 72.4 & 36.2 & 70.6 & 82.1\\
    $FL_{act}$ & 65.4 & \textbf{19.0} & 72.4 & \textbf{32.7} & \textbf{70.9} & \textbf{82.4}\\
    \bottomrule
  \end{tabular}
  \label{tab:ablation-loss}}
  \vspace{-2mm}
\end{table}
\vspace{-4mm}

\subsection{Ablations}
\label{exp:ablation}
\vspace{-1.5mm}
We perform extensive ablation studies to analyze the impact of components of ActionVOS. All ablation experiments are conducted on VISOR dataset and based on ReferFormer-ResNet101. 

\noindent{\bf Classification Head.}
As illustrated in \cref{sec:model}, there is an extra classification head in ActionVOS model to predict objects' positivity. We compare the segmentation results with and without this classification head, and we also test different threshold $\theta$ of this binary classification during inference (\cref{eq:inference}). As shown in \cref{tab:ablation-classification}, using the classification head brings improvements to all metrics, indicating that the head has a strong ability to distinguish active objects, which is a simple yet efficient modification to the model. 
For the threshold $\theta$, using a higher threshold significantly reduce the mis-segmentation of negative objects (decrease in n-mIoU and n-cIoU) while ensuring there is only a slight decrease in positive IoUs (p-mIoU and p-cIoU). 
Considering best trade-off between positive and negative IoUs, we set $\theta=0.75$ for all the experiments. 

\noindent{\bf Text Prompts.}
We compare three types of language prompts, as the design of language prompts is important in language and vision-language tasks \cite{shin2020autoprompt,radford2021learning,zhou2022learning,miao2024improving,tateno2024learning}. 
These three types are as follows:
\begin{itemize}
\vspace{-2mm}
  \item \textit{NoAction}. The text prompt is the object class name. \eg, ``knife''.
  \item \textit{+,Action}. The text prompt is the object class name and the action narration combined with a comma. \eg, ``knife, cut apple''. 
  \item \textit{+sAction}. The text prompt is a natrual sentence consisting of the object class name and action narration, \eg, ``knife used in the action of cut apple''.
\end{itemize}
\vspace{-2.5mm}
We compare these three types with a text decoder frozen and tuned, respectively. As \cref{tab:ablation-prompt} shows, no matter if the text encoder is tuned, \textit{+sAction} enhances the segmentation results most.
This indicates that using a natural language description helps the segmentation model better understand the action. 
In other experiments, the text encoder is tuned and the language prompt is \textit{+sAction}.

\noindent{\bf Action-guided Focal Loss.}
In \cref{sec:loss}, the action-guided focal loss $FL_{act}$ is proposed to reduce the impact of false positives, prioritize truly active objects.
Here, we compare the proposed $FL_{act}$ with focal loss $FL$ \cite{lin2017focal}.
As can be seen in \cref{tab:ablation-loss}, the proposed loss function offers improvements to the gIoU and Acc, and decrease in n-mIoU and n-cIoU. This indicates that the impact of false positives have been reduced.
Visualization in \cref{fig:result-loss} shows the segmentation results when there are objects in both hands. 
The model trained with the proposed action-guided focal loss well prioritizes truly active objects and ignores those false positives.  
In the ``put olive oil'' example, a bottle of olive oil is in the left hand while a bottle of salt is in the right hand. The model trained without action-guided focal loss failed to segment the olive oil and predicted the salt as positive. In contrast, the model trained with our proposed loss successfully segments the olive oil as the only active object.

\section{Conclusion}
\label{sec:conclusion}
In this paper, we propose ActionVOS, a novel action-aware setting for referring video object segmentation. This setting segments active objects in egocentric videos by employing action narrations as an additional language prompt. Specifically, we develop an action-aware labeling module and an action-guided focal loss for ActionVOS. This design enables ActionVOS models to segment active objects with existing readily-available annotations. As for future work, we consider extending ActionVOS by incorporating various action-object relations, reducing the heavy reliance on the availability of dense annotations, and adapting ActionVOS in open-world applications.

%
%

\end{document}